\documentclass[runningheads]{llncs}

\usepackage[T1]{fontenc}
\usepackage{graphicx,verbatim}
\usepackage{cite}
\usepackage{cite}
\usepackage{multirow}
\usepackage{subfigure}
\def\SAM{\text{AutoRad-Lung}}
\usepackage{amsmath}
\usepackage{amsfonts}
\usepackage{bm}

\begin{document}
\title{$\SAM$: A Radiomic-Guided Prompting Autoregressive Vision-Language Model for Lung Nodule Malignancy Prediction}

\author{Sadaf Khademi\inst{1} \and
 Mehran Shabanpour\inst{1} \and
Reza Taleei\inst{2} \and Anastasia Oikonomou\inst{3} \and Arash Mohammadi\inst{1}\thanks{This work was partially supported by Natural Sciences and Engineering Research Council (NSERC) of Canada through NSERC Discovery Grant RGPIN-2023-05654}}
\authorrunning{S. Khademi et al.}
%
\institute{Concordia University, Montreal, Canada
 \and Thomas Jefferson University Hospital, Philadelphia, United States \and
Sunnybrook Health Sciences Centre, Toronto, Canada}

\titlerunning{The AutoRad-Lung}
    
\maketitle              
\begin{abstract}
Lung cancer remains one of the leading causes of cancer-related mortality worldwide. A crucial challenge for early diagnosis is differentiating uncertain cases with similar visual characteristics and closely annotation scores. In clinical practice, radiologists rely on quantitative, hand-crafted Radiomic features extracted from Computed Tomography (CT) images, while recent research has primarily focused on deep learning solutions. More recently, Vision-Language Models (VLMs), particularly Contrastive Language-Image Pre-Training (CLIP)-based models, have gained attention for their ability to integrate textual knowledge into lung cancer diagnosis. While CLIP-Lung models have shown promising results, we identified the following potential limitations: (a) dependence on radiologists’ annotated  attributes, which are inherently subjective and error-prone, (b) use of textual information only during training, limiting direct applicability at inference, and (c) Convolutional-based vision encoder with randomly initialized weights, which disregards prior knowledge. To address these limitations, we introduce AutoRad-Lung, which couples an autoregressively pre-trained VLM, with prompts generated from hand-crafted Radiomics. AutoRad-Lung uses the vision encoder of the Large-Scale Autoregressive Image Model (AIMv2), pre-trained using a multi-modal autoregressive objective. Given that lung tumors are typically small, irregularly shaped, and visually similar to healthy tissue, AutoRad-Lung offers significant advantages over its CLIP-based counterparts by capturing pixel-level differences. Additionally, we introduce conditional context optimization, which dynamically generates context-specific prompts based on input Radiomics, improving cross-modal alignment. Experimental results on the benchmark LIDC-IDRI dataset demonstrate the superiority of benchmark, achieving a relative 6\% accuracy improvement, and relative gains of 16\% in recall and 24\% in F1 score for the ``unsure'' class.

\end{abstract}

\section{Introduction} 
Lung cancer, specifically, Non-small Cell Lung Cancer, remains a leading cause of cancer-related mortality globally~\cite{miller2022cancer}. The diagnosis of pulmonary nodules is of significant importance in the early stages of lung cancer, as timely intervention can dramatically increase the success rate of treatments and improve patient survival. For the early diagnosis of lung cancer, volumetric Computed Tomography (CT) scans offer detailed cross-sectional visual data.
However, complex nature of lung nodules, variability in patient anatomy, and disease progression present significant challenges for current diagnostic models. 
%
Traditionally, lung cancer diagnosis relies on extracting Radiomic features from CT scans~\cite{zhang2017radiomics, rizzo2018radiomics}. These quantitative, hand-crafted features offer valuable insights into the tumor’s morphology, texture, and spatial patterns. The conventional Radiomics pipeline involves extracting thousands of radiomic features, which are subsequently refined through feature selection and reduction techniques to develop a ``Radiomics Signature.'' This signature encapsulates tumor heterogeneity, as well as morphological, textural, and spatial characteristics.

Despite the widespread adoption of conventional hand-crafted Radiomics workflows in clinical practice, advancements in Artificial Intelligence (AI) have led to a paradigm shift toward Deep Learning (DL)-based Radiomics. Consequently, DL-based models for lung nodule classification have been extensively explored~\cite{binczyk2021radiomics, davri2023deep, afshar2019handcrafted}, leveraging diverse architectures such as Convolutional Neural Networks (CNNs)~\cite{dutande2021lncds}, Capsule Networks~\cite{afshar20203d}, and Transformer networks~\cite{azad2024advances}. A further breakthrough in the field has been driven by the emergence of Vision-Language Models (VLMs)~\cite{zhang2024vision}, inspired by advancements in Large Language Models (LLMs)~\cite{brown2020language}. The integration of language with visual representations through VLMs is anticipated to be a key driver of progress in medical image analysis~\cite{bordes2024}. In particular, recent research has focused on the efficient adaptation of the Contrastive Language-Image Pre-Training (CLIP) model~\cite{radford2021} for lung nodule classification~\cite{lei2023, sun2024}. 

While CLIP-based models for lung diagnosis~\cite{lei2023, sun2024} have made significant progress, they still face key limitations. First, these models depend on text attributes annotated by radiologists, whose analyses of CT images are inherently subjective and prone to errors. Moreover, since these annotations are only utilized during training, they offer no direct applicability during the inference phase. Second, the vision encoder backbone used in CLIP-Lung~\cite{lei2023} and Nodule-CLIP~\cite{sun2024} is based on ResNet with randomly initialized weights. This approach contradicts the fundamental principle of contrastive pre-training, as randomized initialization disregards prior knowledge that could enhance feature extraction. Furthermore, CNN-based models like ResNet rely on local receptive fields, limiting their ability to capture long-range dependencies. This is particularly problematic in 3D CT scans, where global context is critical for accurate diagnosis.

To overcome the aforementioned challenges, we propose the $\SAM$ architecture, which integrates multi-modal autoregressive pre-training~\cite{el2024, yang2019xlnet} with hand-crafted Radiomics. While the autoregressive vision encoder captures deep image features, Radiomic features contextualized as textual description provide domain-specific, handcrafted information. Combining deep autoregressively extracted features with Radiomics enables the model to leverage both local pixel-level details and global shape/texture patterns, ultimately improving diagnostic accuracy and clinical relevance. To encode visual information, we use the vision encoder of the recently introduced Large-Scale Autoregressive Image Model (AIMv2)~\cite{fini2024}. AIMv2 is pre-trained using a multi-modal autoregressive objective that reconstructs image patches and text tokens. Given that lung tumors are typically small, irregularly shaped, and visually similar to healthy tissue, AIMv2 offers significant advantages over its CLIP-based counterparts by capturing pixel-level differences. The pixel-level reconstruction loss ensures that the vision encoder learns fine-grained spatial features, which are essential for detecting small tumors and early-stage abnormalities. To further improve cross-modal alignment, we employ conditional context optimization and prompt tuning, dynamically generating context-specific prompts based on input Radiomic features. Unlike static prompts that apply the same text input across all cases, our approach tailors each prompt to the specific Radiomic context. More specifically, use of prompt tuning allows the text encoder to generate context-specific prompts that reflect the clinical characteristics of each tumor, improving cross-modal alignment between image patches and text tokens. In summary, the contributions of this paper are summarized as follows:
\begin{itemize}
\item We introduce $\SAM$, which, to the best of our knowledge, is the first model to couple an autoregressively pre-trained VLM, AIMv2, with prompts generated from hand-crafted Radiomics. $\SAM$ eliminates dependence on radiologist-annotated attributes, leverages dense supervision, and captures fine-grained spatial features.
\item We design a context optimization strategy that dynamically generates context-specific prompts based on input Radiomics. This allows integration of textual input both at training and inference. 
\item $\SAM$ demonstrates improved sensitivity in detecting uncertain cases, which is particularly crucial in medical applications.
\end{itemize}

\section{Methodology}

\subsection{Problem Formulation}
We consider the problem of lung nodule classification based on a dataset constructed as $\bm{\mathcal{D}} = \{\mathcal{I}, \mathcal{Y}, \mathcal{C}, \mathcal{R}\}$, where $\mathcal{I} = \{\bm{I}_j\}_{j=1}^{N_i}$ is the input set of $N_i$ CT images. Set $\mathcal{Y} = \{y_j\}_{j=1}^{N_i}$ consists of $N_c$ nodule classification labels for ($1 \leq y_j \leq N_c$). Set $\mathcal{C} = \{c_k\}_{k=1}^{N_c}$ is text embeddings associated with the $N_c$ nodule classes. Finally, $\mathcal{R} = \{\bm{r}_j\}_{j=1}^{N_i}$ is the set of Radiomics extracted from each image,  $\bm{I}_j$, where  $\bm{r}_j \in \mathbb{R}^{N_r \times 1}$ with $N_r$ being the dimension of the extracted Radiomics vector. The task at hand is to learn a context prompt that guides the model based on hand-crafted nodule characteristics, enabling it to identify ambiguous nodules that require further clinical review.

\subsection{The $\SAM$ Framework}
\begin{figure}[t!]
\includegraphics[width=\textwidth]{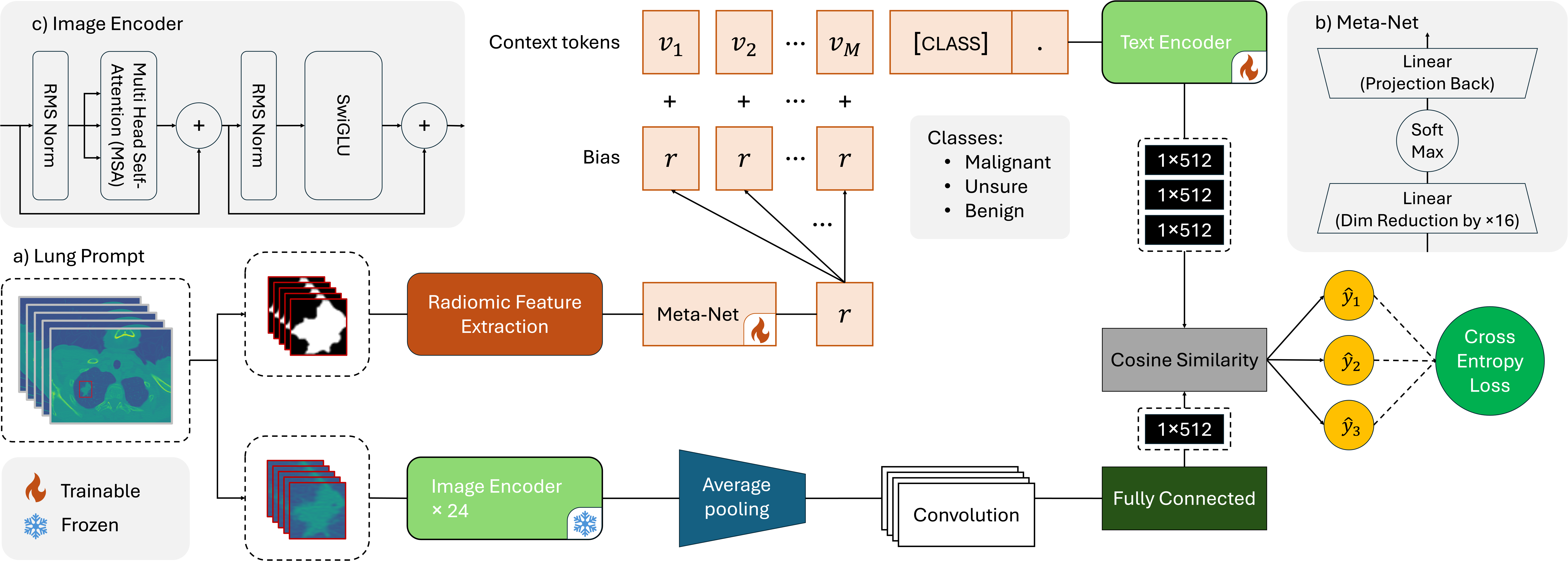}
\caption{The proposed \SAM\ framework consisting of two parallel paths, one for extracting hand-crafted Radiomics to guide the model in learning context tokens, and the other for extracting deep features, which are then aligned.} \label{fw}
\end{figure}
Fig.~\ref{fw} illustrates the $\SAM$'s architecture, which consists of the following key components: (1) An autoregressive image encoder $f(\cdot)$; (2) Radiomic feature extraction module; (3) Meta-Network (MetaNet), which transforms the extracted Radiomics into latent representation $h_{\bm{\theta}}(\cdot)$; (4) A text encoder $g_{\bm{\eta}}(\cdot)$, and; (5) A fusion mechanism to align information obtained from each of the two underlying encoders.
The processing pipeline begins with extracting features from cropped nodule slices. These images are then passed through the vision encoder, generating high-level image features. The features extracted from individual slices within a nodule's volume are combined using an average pooling layer to create a unified visual representation of the nodule. Simultaneously, Radiomic features are extracted from the masked nodule in the middle slice of each volume, capturing key intensity, texture, and shape-based attributes. These Radiomics are then fed into the MetaNet, to generate a conditional token (vector) for each input, which is then combined with the context vectors. To enhance contextual understanding, a randomly initialized $50$-dimensional context embedding is added to this transformed feature vector, which is subsequently processed by the text encoder. The final representations from the image encoder and text encoder are aligned through the fusion module using cosine similarity. Alignment logits are then used to compute the cross-entropy loss, which guides the optimization of the MetaNet and context embeddings.

\subsection{Radiomic Features of Lung Nodule} 
To address reliance on attribute annotations from radiologists, we perform Radiomic feature extraction for baseline screen-detected pulmonary nodules. For this purpose PyRadiomics (version $3.0.1$) is used. Due to heterogeneity in image acquisition settings, all images and masks were resampled and interpolated to have unit (1mm$^3$) voxel spacing. We used a linear interpolator for images and nearest-neighbours interpolator for masks (to preserve labels). Grayscale intensities were discretized into bins using a width of $25$ for histogram-based features. Voxel intensities were right-shifted by $1,000$ units prior to feature extraction to avoid negative values during feature computations. Overall, the following Radiomic features classes are considered: (1) First-order statistics, (2) Shape-based features, (3) Gray level co-occurrence matrix; (4) Gray level run length matrix, (5) Gray level size zone matrix, (6) Neighbouring gray tone difference matrix, and (7) Gray level dependence matrix. Shape and intensity-based features are extracted using the original image, while intensity-based features are extracted from images after applying different transformations, i.e., wavelet, Laplacian of Gaussian (LoG), Square, SquareRoot, Logarithm, Exponential, Gradient, and Local-Binary-Pattern-2D. In total, $1,500$ hand-crafted Radiomics are extracted per nodule~\cite{warkentin2024radiomics}.

\subsection{Textual Encoding} 
To encode textual information, we used the text encoder from the OpenAI CLIP model~\cite{radford2021}. The CLIP text encoder is inspired by the GPT-2 design and is optimized to encode input text into a shared text-image embedding space. The utilized architecture comprises of twelve residual attention blocks positioned in a sequential fashion. Each residual attention block consists of a multi-head self-attention module with eight head for capturing token dependencies.

\subsection{Visual Encoding} 
As stated previously, we use AIMv2 (particularly, the pre-trained AIMv2-large-patch$14-224$) as the vision encoder of the $\SAM$. Autoregressive modeling has a long history in statistics, however, recently, they have emerged as an efficient pre-training approach~\cite{el2024}. Intuitively speaking, the idea is to model text tokens and image patches as a unified sequence and perform sequential prediction via a causal multi-modal decoder. The output of the decoder is then used to predict the next tokens in each modality, separately via a dedicated linear head. More specifically, each input image $\bm{I}_j$, for ($1 \leq j \leq N_i$) is partitioned into $I$ non-overlapping patches $\bm{I}^{(i)}_j$, for ($1 \leq i \leq I$), which form a sequence of tokens. 
The text sequence is similarly reduced into its constituent sub-words, which are then concatenated with image tokens. This allows construction of attention maps between image and text tokens. The image information is purposely placed first allowing the visual context to fully inform the text generation process. This creates one specific and continuous sequence, which is processed to predict the next element in the sequence. Such a sequential structure with image-first positioning results in a totally different vision encoder compared to that of the CLIP where image and text tokens are processed in parallel. In AIMv2, image and text are jointly predicted and reconstructed while in CLIP the contrastive training approach processes text and image modalities separately and then targets alignment of existing image and text pairs. 

\subsection{Meta-Net Conditional Context Optimization} 
The concept of Context Optimization (CoOp)~\cite{zhou2022learning} originates from efforts to enhance the adaptability of pre-trained VLMs. CoOp tackles inefficiencies in manual prompt design (e.g., ``a nodule image of malignant/benign'') by employing $M$ learnable context vectors, $\{\bm{v}_1, \dots, \bm{v}_M\}$, each matching the dimensionality of word embeddings. These vectors, shared across classes, combine with class-specific embeddings $\bm{c}_i$ to form prompts $t_i = [\bm{v}_1, \dots, \bm{v}_M, \bm{c}_i]$, optimized end-to-end for downstream tasks. However, CoOp’s static prompts limit generalization to unseen classes within the same task. To address this, Conditional Context Optimization (CoCoOp)~\cite{zhou2022conditional} extends CoOp by integrating a lightweight neural network, termed Meta-Net ($h_{\bm{\theta}}(\cdot)$), which generates an input-specific token $\delta = h_{\bm{\theta}}(x)$ for each image. This modifies context vectors dynamically as $\bm{v}_m(x) = \bm{v}_m + \delta$, resulting in instance-conditioned prompts $t_i(x) = [\bm{v}_1(x), \dots, \bm{v}_M(x), \bm{c}_i]$. This adaptability reduces overfitting to specific classes, focusing instead on the entire task. The prediction probability is:
\begin{eqnarray}
p(y|x) = \frac{\exp(\text{sim}(x, g_{\bm{\eta}}(t_y(x))) / \tau)}{\sum_{i=1}^{N_c} \exp(\text{sim}(x, g_{\bm{\eta}}(t_i(x))) / \tau)},
\end{eqnarray}
where $\text{sim}(\cdot, \cdot)$ measures similarity, $g_{\bm{\eta}}(\cdot)$ is the text encoder,  $\tau$ is a temperature parameter, and $N_c$ is number of classes. During training, CoCoOp jointly updates the context vectors ${\bm{v}_m}$ and Meta-Net parameters $\bm{\theta}$. The Meta-Net is a two-layer bottleneck architecture (Linear-ReLU-Linear).

\section{Experiments}
\begin{figure}[t!]
\centering
\mbox{\subfigure[]{\includegraphics[scale=0.4]{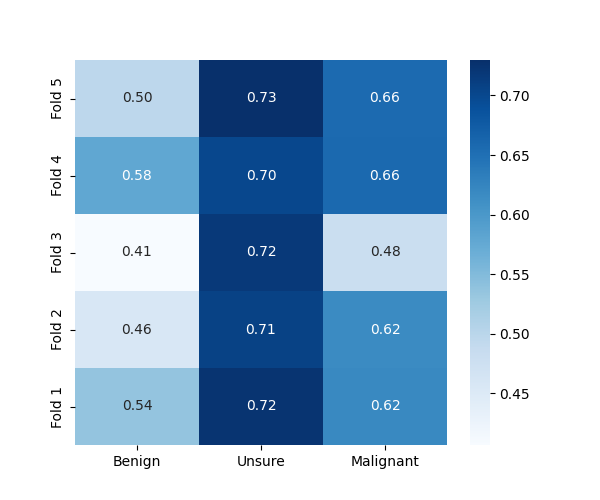}}
\hspace{-.6in}
\subfigure[]{\includegraphics[scale=0.4]{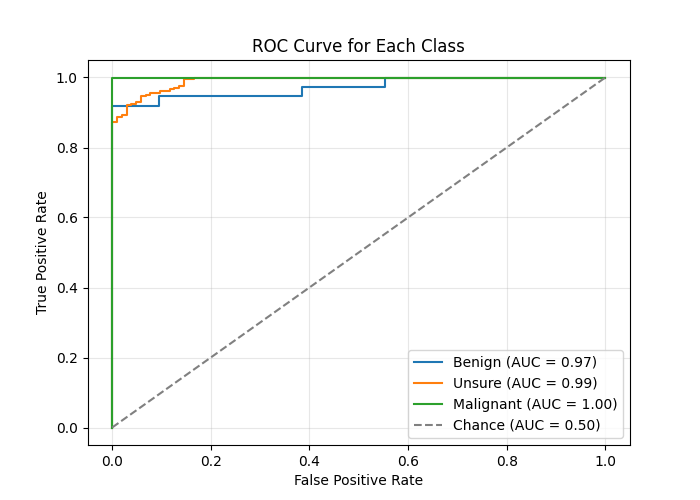}}}
\caption{(a) Heatmap of F1-scores achieved for each class across $5$ folds. (b) Receiver Operating Characteristic (ROC) curve for each class in Fold $1$. \label{Fig:1}}
\end{figure}
\subsection{Dataset}
We use the benchmark Lung Image Database Consortium (LIDC) and Image Database Resource Initiative (IDRI) dataset~\cite{armato2011lung}, which focuses on the pulmonary nodule detection/classification task based on low-dose CT images. The LIDC-IDRI dataset consists of CT volumes from $1,010$ patients and includes nodule annotation scores (ranging from $1-5$) related to its malignancy progression.  Following prior works~\cite{lei2023, wu2019learning},  we consider a three class problem consisting of benign, malignant, and unsure categories.  A nodule is categorized as ``benign'' when its average score is less than $2.5$, while a nodule with an average score greater than $3.5$ is considered ``malignant''. Finally, a nodule with an average score of $2.5$ to $3.5$ is categorized as ``unsure''.

\subsection{Pre-processing}
Following~\cite{lei2023}, CT scans and their corresponding annotations are pre-processed separately to facilitate context extraction of pulmonary nodules. This is performed to ensure each nodule has at least one radiologist-provided annotation. Initially, all slices containing annotations are identified for a given nodule. Then, the endpoints of each series, i.e., the first and last slices of the series where the nodule is visible, are identified. Finally, the middle slice between the identified endpoints is selected as a representative for Radiomics analysis. To generate a nodule mask, we aggregate annotations based on radiologist consensus, defining the nodule boundary only if at least $50$\% of the annotators agree. This consensus-based mask, along with the selected central slice, forms the input for subsequent Radiomics feature extraction. Additionally, to standardize input dimensions, we crop each nodule-containing slice into a square region, with a side length twice the nodule's equivalent diameter, centered on the annotated centroid. The cropped region is then resized to a fixed resolution for compatibility with the vision encoder.

\subsection{Experimental Design}
\begin{table*}[t!]
\setlength{\tabcolsep}{5pt}
\centering
\caption{Classification results on LIDC-IDRI dataset.}
\begin{tabular}{l|c|c|c|c|c|c|c}
\hline
\multirow{2}{*}{\textbf{Method}} & \textbf{Accuracy} & \multicolumn{2}{|c|}{\textbf{Benign}} & \multicolumn{2}{c}{\textbf{Malignant}} & \multicolumn{2}{|c}{\textbf{Unsure}} \\
\cline{3-8}
 & (\%) & Recall & F1 & Recall & F1 & Recall & F1 \\
\hline
ResNet18~\cite{lei2023} & 54.2 $\pm$ 0.6 & 72.2 & 62.0 & 64.4 & 61.3 & 29.0 & 36.6 \\
UDM~\cite{wu2019learning}  & 54.6 $\pm$ 0.4 & 76.7 & 64.3 & 49.5 & 53.5 & 32.5 & 39.5 \\
CLIP~\cite{radford2021}  & 56.6 $\pm$ 0.3 & 59.5 & 59.2 & 55.2 & 60.0 & 53.9 & 52.2 \\
CoCoOp~\cite{zhou2022conditional}  & 56.8 $\pm$ 0.6 & 59.0 & 59.2 & 55.2 & 60.0 & 55.1 & 52.8 \\
AIMv2~\cite{fini2024} & 58.5 $\pm$ 0.3 & 62.5 & 55.2 & 43.6 & 45.6 & 51.3 & 52.3 \\
CLIP-Lung~\cite{lei2023} & 60.9 $\pm$ 0.4 & 67.5 & \textbf{64.4} & 60.9 & \textbf{66.3} & 53.4 & 54.1 \\
\textbf{\SAM} & \textbf{64.6} $\pm$ 1.7 & \textbf{75.3} & 49.5 & \textbf{65.6} & 60.6 & \textbf{62.3} & \textbf{71.61} \\
\hline
\end{tabular}
\label{tab:lidc_a_results}
\end{table*}
The $\SAM$ is trained with a learning rate of $0.0001$ following the cosine decay and stochastic gradient descent as its optimizer with momentum $0.9$ and weight decay $5e-7$.
All of our experiments are implemented with PyTorch and trained with NVIDIA GeForce RTX $3090$. The number of context tokens in learnable prompt is set to $50$. The experimental results are reported as average values through stratified $5$-fold cross-validation.

\subsection{Results and Performance Analysis}
While much of the literature has focused on binary classification, in this study, our goal is to highlight the importance of unsure data, a category that has been largely overlooked. Incautious predictions for unsure cases at early-stages can result in irreversible consequences, such as missing critical treatment opportunities. We emphasize the need to effectively address these uncertain cases. 
According to the results presented in Table~\ref{tab:lidc_a_results}, the effect of the prompt embedding on classification accuracy is evident across the two main backbone architectures, ResNet$18$ and AIMv2. This suggests that a well pre-trained image encoder significantly improves classification performance. The notable gap (approximately $6$\%) between each backbone and its VLM design further highlights the contribution of the text encoder. Additionally, our model outperforms the baseline study in both accuracy and sensitivity. Our primary objective (enhancing classification performance for the ``unsure'' class) is effectively demonstrated in Fig.~\ref{Fig:1}(a) based on F1-score values.
Moreover, as can be observed in Fig.~\ref{Fig:1}(b), high AUC values, obtained using the one-vs-rest (OvR) strategy, indicate that the model effectively differentiates each class from the rest, consistently ranking the correct class higher than incorrect ones. This suggests that the model makes confident and well-calibrated predictions, demonstrating strong discriminative ability in this multi-class setting. Finally, we examine the impact of the number of context tokens, and as illustrated in Table~\ref{tab:Context}, accuracy improves up to a certain threshold before it starts to decline. This suggests that while increasing the number of context tokens, initially, enhances feature learning, exceeding an optimal number may lead to overfitting or redundancy, reducing classification performance.

\begin{table*}[t!]
\centering
\caption{Model performance in fold 1 across different numbers of context tokens.}
\begin{tabular}{l|c|c|c|c|c|c|c}
\hline
\multirow{2}{*}{\textbf{Method}} & \multicolumn{7}{|c}{\textbf{Number of context tokens }} \\

 & $10$ & $20$ & $30$ & $40$ & \textbf{50} & $60$ & $70$ \\ 
\hline
Accuracy (\%)& $63.27$ & $63.65$ & $63.84$ & $61.95$ & \textbf{66.10} & $65.34$ & $62.33$ \\
\hline
\end{tabular}
\label{tab:Context}
\end{table*}
\section{Conclusion}
Pulmonary nodule prediction faces challenges in labeling uncertain cases with similar visual characteristics and close annotation scores, particularly in early stages. In this context, we proposed the AutoRad-Lung, a Radiomics-guided framework built over an autoregressively pre-trained VLM. AutoRad-Lung eliminates dependence on radiologist-annotated attributes, leverages dense supervision, and captures fine-grained spatial features. Radiomics features, extracted from the masked nodule’s middle slice, encode key intensity, texture, and shape attributes. These features are processed by MetaNet to generate conditional tokens, which are integrated with context embeddings and refined through a fusion module aligning image and text representations. Our findings underscore the value of Radiomics features in enhancing VLM accuracy for lung nodule classification.

\newpage
\bibliographystyle{splncs04}
\bibliography{refs}

\begin{thebibliography}{10}
\providecommand{\url}[1]{\texttt{#1}}
\providecommand{\urlprefix}{URL }
\providecommand{\doi}[1]{https://doi.org/#1}

\bibitem{afshar2019handcrafted}
Afshar, P., Mohammadi, A., Plataniotis, K.N., Oikonomou, A., Benali, H.: From
  handcrafted to deep-learning-based cancer radiomics: challenges and
  opportunities. IEEE Signal Processing Magazine  \textbf{36}(4),  132--160
  (2019)

\bibitem{afshar20203d}
Afshar, P., Oikonomou, A., Naderkhani, F., Tyrrell, P.N., Plataniotis, K.N.,
  Farahani, K., Mohammadi, A.: 3d-mcn: a 3d multi-scale capsule network for
  lung nodule malignancy prediction. Scientific reports  \textbf{10}(1), ~7948
  (2020)

\bibitem{armato2011lung}
Armato~III, S.G., McLennan, G., Bidaut, L., McNitt-Gray, M.F., Meyer, C.R.,
  Reeves, A.P., Zhao, B., Aberle, D.R., Henschke, C.I., Hoffman, E.A., et~al.:
  The lung image database consortium (lidc) and image database resource
  initiative (idri): a completed reference database of lung nodules on ct
  scans. Medical physics  \textbf{38}(2),  915--931 (2011)

\bibitem{azad2024advances}
Azad, R., Kazerouni, A., Heidari, M., Aghdam, E.K., Molaei, A., Jia, Y., Jose,
  A., Roy, R., Merhof, D.: Advances in medical image analysis with vision
  transformers: a comprehensive review. Medical Image Analysis  \textbf{91},
  103000 (2024)

\bibitem{binczyk2021radiomics}
Binczyk, F., Prazuch, W., Bozek, P., Polanska, J.: Radiomics and artificial
  intelligence in lung cancer screening. Translational lung cancer research
  \textbf{10}(2), ~1186 (2021)

\bibitem{bordes2024}
Bordes, F., Pang, R.Y., Ajay, A., Li, A.C., Bardes, A., Petryk, S., Ma{\~n}as,
  O., Lin, Z., Mahmoud, A., Jayaraman, B., et~al.: An introduction to
  vision-language modeling. arXiv preprint arXiv:2405.17247  (2024)

\bibitem{brown2020language}
Brown, T., Mann, B., Ryder, N., Subbiah, M., Kaplan, J.D., Dhariwal, P.,
  Neelakantan, A., Shyam, P., Sastry, G., Askell, A., et~al.: Language models
  are few-shot learners. Advances in neural information processing systems
  \textbf{33},  1877--1901 (2020)

\bibitem{davri2023deep}
Davri, A., Birbas, E., Kanavos, T., Ntritsos, G., Giannakeas, N., Tzallas,
  A.T., Batistatou, A.: Deep learning for lung cancer diagnosis, prognosis and
  prediction using histological and cytological images: a systematic review.
  Cancers  \textbf{15}(15), ~3981 (2023)

\bibitem{dutande2021lncds}
Dutande, P., Baid, U., Talbar, S.: Lncds: A 2d-3d cascaded cnn approach for
  lung nodule classification, detection and segmentation. Biomedical signal
  processing and control  \textbf{67},  102527 (2021)

\bibitem{el2024}
El-Nouby, A., Klein, M., Zhai, S., Bautista, M.A., Toshev, A., Shankar, V.,
  Susskind, J.M., Joulin, A.: Scalable pre-training of large autoregressive
  image models. arXiv preprint arXiv:2401.08541  (2024)

\bibitem{fini2024}
Fini, E., Shukor, M., Li, X., Dufter, P., Klein, M., Haldimann, D., Aitharaju,
  S., da~Costa, V.G.T., B{\'e}thune, L., Gan, Z., et~al.: Multimodal
  autoregressive pre-training of large vision encoders. arXiv preprint
  arXiv:2411.14402  (2024)

\bibitem{lei2023}
Lei, Y., Li, Z., Shen, Y., Zhang, J., Shan, H.: Clip-lung: Textual
  knowledge-guided lung nodule malignancy prediction. In: International
  Conference on Medical Image Computing and Computer-Assisted Intervention. pp.
  403--412. Springer (2023)

\bibitem{miller2022cancer}
Miller, K.D., Nogueira, L., Devasia, T., Mariotto, A.B., Yabroff, K.R., Jemal,
  A., Kramer, J., Siegel, R.L.: Cancer treatment and survivorship statistics,
  2022. CA: a cancer journal for clinicians  \textbf{72}(5),  409--436 (2022)

\bibitem{radford2021}
Radford, A., Kim, J.W., Hallacy, C., Ramesh, A., Goh, G., Agarwal, S., Sastry,
  G., Askell, A., Mishkin, P., Clark, J., et~al.: Learning transferable visual
  models from natural language supervision. In: International conference on
  machine learning. pp. 8748--8763. PmLR (2021)

\bibitem{rizzo2018radiomics}
Rizzo, S., Botta, F., Raimondi, S., Origgi, D., Fanciullo, C., Morganti, A.G.,
  Bellomi, M.: Radiomics: the facts and the challenges of image analysis.
  European radiology experimental  \textbf{2}, ~1--8 (2018)

\bibitem{sun2024}
Sun, L., Zhang, M., Lu, Y., Zhu, W., Yi, Y., Yan, F.: Nodule-clip: Lung nodule
  classification based on multi-modal contrastive learning. Computers in
  Biology and Medicine  \textbf{175},  108505 (2024)

\bibitem{warkentin2024radiomics}
Warkentin, M.T., Al-Sawaihey, H., Lam, S., Liu, G., Diergaarde, B., Yuan, J.M.,
  Wilson, D.O., Atkar-Khattra, S., Grant, B., Brhane, Y., et~al.: Radiomics
  analysis to predict pulmonary nodule malignancy using machine learning
  approaches. Thorax  \textbf{79}(4),  307--315 (2024)

\bibitem{wu2019learning}
Wu, B., Sun, X., Hu, L., Wang, Y.: Learning with unsure data for medical image
  diagnosis. In: Proceedings of the IEEE/CVF International Conference on
  Computer Vision. pp. 10590--10599 (2019)

\bibitem{yang2019xlnet}
Yang, Z., Dai, Z., Yang, Y., Carbonell, J., Salakhutdinov, R.R., Le, Q.V.:
  Xlnet: Generalized autoregressive pretraining for language understanding.
  Advances in neural information processing systems  \textbf{32} (2019)

\bibitem{zhang2024vision}
Zhang, J., Huang, J., Jin, S., Lu, S.: Vision-language models for vision tasks:
  A survey. IEEE Transactions on Pattern Analysis and Machine Intelligence
  (2024)

\bibitem{zhang2017radiomics}
Zhang, Y., Oikonomou, A., Wong, A., Haider, M.A., Khalvati, F.: Radiomics-based
  prognosis analysis for non-small cell lung cancer. Scientific reports
  \textbf{7}(1),  46349 (2017)

\bibitem{zhou2022conditional}
Zhou, K., Yang, J., Loy, C.C., Liu, Z.: Conditional prompt learning for
  vision-language models. In: Proceedings of the IEEE/CVF conference on
  computer vision and pattern recognition. pp. 16816--16825 (2022)

\bibitem{zhou2022learning}
Zhou, K., Yang, J., Loy, C.C., Liu, Z.: Learning to prompt for vision-language
  models. In: Proceedings of the IEEE/CVF conference on computer vision and
  pattern recognition. pp. 2337--2348 (2022)

\end{thebibliography}
\end{document}